\documentclass[10pt,a4paper,twocolumn]{article}

\usepackage[utf8]{inputenc}
\usepackage[T1]{fontenc}
\usepackage{lmodern}
\usepackage[margin=0.75in]{geometry}
\usepackage{amsmath,amssymb}
\usepackage{booktabs}
\usepackage{graphicx}
\usepackage{xcolor}
\usepackage{url}
\usepackage{natbib}
\usepackage{hyperref}  

\hypersetup{
    colorlinks=true,
    linkcolor=blue!70!black,
    citecolor=green!50!black,
    urlcolor=blue!70!black,
}

\title{Federated Few-Shot Learning on Neuromorphic Hardware:\\
An Empirical Study Across Physical Edge Nodes}

\author{
Steven Motta \and Gioele Nanni
}

\date{March 13, 2026}

\begin{document}
\maketitle

\begin{abstract}
Federated learning on neuromorphic hardware remains unexplored because on-chip spike-timing-dependent plasticity (STDP) produces binary weight updates rather than the floating-point gradients assumed by standard algorithms. We build a two-node federated system with BrainChip Akida AKD1000 processors and run approximately 1{,}580 experimental trials across seven analysis phases. Of four weight-exchange strategies tested, neuron-level concatenation (FedUnion) consistently preserves accuracy while element-wise weight averaging (FedAvg) destroys it ($p = 0.002$). Domain-adaptive fine-tuning of the upstream feature extractor accounts for most of the accuracy gains, confirming feature quality as the dominant factor. Scaling feature dimensionality from 64 to 256 yields $77.0\%$ best-strategy federated accuracy ($n=30$, $p < 0.001$). Two independent asymmetries (wider features help federation more than individual learning, while binarization hurts federation more) point to a shared \emph{prototype complementarity} mechanism: cross-node transfer scales with the distinctiveness of neuron prototypes.
\end{abstract}

\section{Introduction}
\label{sec:intro}

Neuromorphic processors such as the BrainChip Akida AKD1000~\citep{brainchip2023} execute spiking neural network (SNN)~\citep{maass1997networks,roy2019towards} inference and on-chip learning at milliwatt power budgets, making them attractive for edge deployment~\citep{shi2016edge}. Their native learning mechanism, spike-timing-dependent plasticity (STDP)~\citep{bi1998synaptic}, produces sparse binary weight updates rather than the floating-point gradients that federated learning algorithms~\citep{mcmahan2017communication,kairouz2021advances} expect. No prior work has attempted to federate STDP-learned models across physical neuromorphic devices: existing studies operate entirely in simulation and rely on surrogate-gradient training instead of on-chip plasticity.

We build a two-node federated system with Akida AKD1000 coprocessors on Raspberry Pi~5 boards and run approximately 1{,}580~experimental trials across seven analysis phases. Our contributions are:

\begin{enumerate}
    \item A \textbf{two-stage architecture} that works around the Akida IP version mismatch (v2 feature extractor, v1 STDP edge learner) together with an \textbf{on-device fine-tuning pipeline} using quantization-aware training~\citep{jacob2018quantization} on commodity Raspberry Pi hardware.

    \item \textbf{Four federation strategies} for STDP weight matrices. Neuron-level concatenation (FedUnion) preserves accuracy across all tested conditions; element-wise averaging (FedAvg) consistently destroys the sparse selectivity patterns that STDP produces.

    \item An \textbf{experimental framework} covering 42~STDP configurations $\times$ up to 30~trials $\times$ seven analysis phases (${\sim}1{,}580$ runs total), with seeded splits, software baselines, bootstrap CIs, and Cohen's $d$ effect sizes.

    \item \textbf{Feature dimensionality scaling} from 64 to 256 dimensions, and a \textbf{prototype complementarity} mechanism that explains two observed asymmetries: wider features amplify the federation gain while binarization amplifies the federation penalty.
\end{enumerate}

\section{Related Work}
\label{sec:related}

\paragraph{Neuromorphic Computing and SNNs.}
Spiking neural networks (SNNs)~\citep{maass1997networks} communicate via discrete spikes rather than continuous activations, enabling event-driven computation at milliwatt-scale power~\citep{roy2019towards,pfeiffer2018deep}. Hardware platforms include Intel Loihi~\citep{davies2018loihi}, a research manycore chip with programmable learning rules, and IBM TrueNorth~\citep{merolla2014truenorth}, which integrated one million neurons on a single die. The BrainChip Akida AKD1000~\citep{brainchip2023} differs from these research platforms in that it ships as a standard PCIe peripheral and supports on-chip STDP-based learning~\citep{bi1998synaptic} alongside inference. Akida has been applied to single-device keyword spotting~\citep{brainchip2023,ds_cnn_kws} and visual classification~\citep{akida_models}; no multi-device or federated deployments have been reported.

\paragraph{Federated Learning.}
Federated Averaging (FedAvg)~\citep{mcmahan2017communication} trains models across decentralized nodes without sharing raw data. Subsequent work has tackled non-IID distributions~\citep{li2020federated,zhao2018federated}, communication cost~\citep{konecny2016federated}, device heterogeneity~\citep{li2020federated_survey}, and numerous open problems surveyed by \citet{kairouz2021advances}. All of these algorithms aggregate floating-point gradient updates, a primitive that has no direct analogue in STDP-based on-chip learning.

\paragraph{Federated Learning on Neuromorphic Hardware.}
Table~\ref{tab:prior_work} summarises the small body of work in this area; all of it is simulation-only. \citet{skatchkovsky2020federated} trained SNNs on MNIST-DVS across two simulated nodes, transmitting only weights above a threshold and showing convergence on par with centralized training at lower communication cost. \citet{yang2022lead} proposed LEAD, a leader-election protocol in which the best-performing node's weights guide aggregation; on SHD and DVS-Gesture with 3+ simulated nodes they reported $94$--$96\%$ accuracy. \citet{venkatesha2021federated} federated SNNs on CIFAR-10 with 5+ simulated clients using surrogate-gradient backpropagation. A common thread is that all three rely on surrogate gradients rather than on-chip STDP, and none deploy on physical hardware. Our experiments are, to our knowledge, the first to exchange STDP-learned weights between physical neuromorphic processors.

\begin{table}[t]
\centering\footnotesize
\caption{Comparison with prior federated neuromorphic learning. P = physical hardware; S = simulated.}
\label{tab:prior_work}
\setlength{\tabcolsep}{3pt}
\begin{tabular}{@{}lcccc@{}}
\toprule
Study & Hw. & $N$ & Task & Acc.\ (\%) \\
\midrule
Skatchkovsky+ \citeyearpar{skatchkovsky2020federated} & S & 2 & MNIST-DVS & ${\sim}90^*$ \\
Yang+ \citeyearpar{yang2022lead} & S & 3+ & SHD/DVS & 94--96 \\
Venkatesha+ \citeyearpar{venkatesha2021federated} & S & 5+ & CIFAR-10 & ${\sim}90^*$ \\
\midrule
This work & P & 2--4 & Speech Cmd. & 77.0 \\
\bottomrule
\multicolumn{5}{@{}l@{}}{\scriptsize $^*$Estimated from figures.}
\end{tabular}
\end{table}

\paragraph{Few-Shot Learning and Transfer Learning.}
Few-shot learning aims to classify new categories from minimal examples~\citep{wang2020generalizing}. Prototypical networks~\citep{snell2017prototypical} learn embedding spaces in which class prototypes are compared via distance functions; STDP's per-neuron weight vectors play an analogous role, as we discuss in \S\ref{sec:federation_asymmetry}. Fine-tuning pretrained feature extractors for new domains~\citep{pan2010survey} is well established for CNNs but unexplored in combination with neuromorphic STDP learners that require binary inputs. Feature binarization has been studied in binary neural networks~\citep{courbariaux2016binarized}, yet how binarization interacts with federated aggregation of STDP weights is an open question that our experiments address directly.

\section{Proposed Method}
\label{sec:method}

\subsection{Two-Stage Pipeline: Bridging IP Versions}
\label{sec:pipeline}

A critical engineering constraint is the Akida SDK's IP version architecture. The pretrained DS-CNN keyword spotting model~\citep{zhang2017hello} converts to Akida v2, but the \texttt{FullyConnected} layer required for STDP edge learning is only available in v1. These versions \emph{cannot be mixed} within a single model.

Our solution is a \textbf{two-stage pipeline}:
\begin{enumerate}
    \item \textbf{Stage~1, Feature Extractor (v2, software):} The DS-CNN~\citep{zhang2017hello,ds_cnn_kws} is converted from Keras to Akida v2, with the classification head removed. In the base configuration, this maps MFCC spectrograms~\citep{davis1980comparison} of shape $(49, 10, 1)$ to 64-dimensional int8 feature vectors. In Phase~2+, the DS-CNN is fine-tuned on target classes before head removal (\S\ref{sec:finetuning}).

    \item \textbf{Stage~2, Edge Learner (v1, hardware):} A standalone v1 model with a single \texttt{FullyConnected} layer compiled with the \texttt{AkidaUnsupervised} STDP optimizer. It accepts 1-bit binary inputs and maps to the AKD1000 for on-chip learning.
\end{enumerate}

\paragraph{Wide feature variants.} For feature dimensionality scaling experiments (\S\ref{sec:wide_results}), the DS-CNN classification head is replaced with a wider projection: a \texttt{QuantizedDense} layer with 128 or 256 output units, followed by a \texttt{Dequantizer}. After \texttt{cnn2snn.convert()}, which stops at the \texttt{Dequantizer} boundary, the resulting Akida model outputs 64-dimensional features as before. The wider projection weights are saved separately as NumPy \texttt{.npy} files, and the final wide features are computed via post-Akida matrix multiplication in NumPy: $\mathbf{f}_{\text{wide}} = \mathbf{f}_{64} \cdot W_{\text{proj}}$. This workaround addresses a \texttt{cnn2snn} limitation where non-quantized layers after the \texttt{Dequantizer} are ignored during conversion.

\textbf{Hardware boundary.} The NumPy projection means that for wide feature experiments, only the 64-dim backbone feature extraction runs through the Akida-converted model (on the ARM CPU, not the AKD1000), and the $64 \to D$ projection is entirely in software. The STDP edge learner still runs on-chip on the AKD1000, so on-chip learning is preserved, but the feature pipeline is increasingly software-dependent. We return to this caveat in \S\ref{sec:limitations}.

\begin{figure*}[t]
\centering
\includegraphics[width=\textwidth,trim=0 220 0 70,clip]{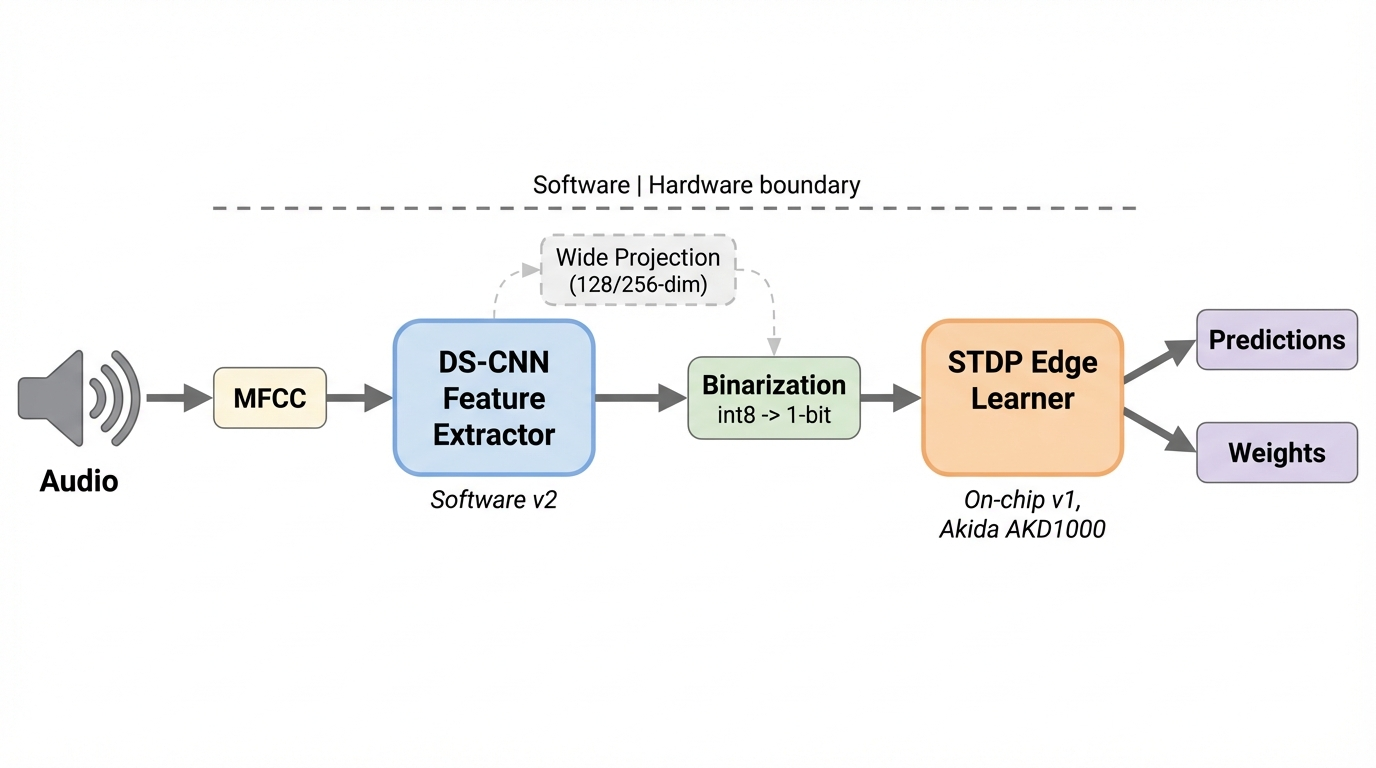}
\caption{Two-stage pipeline. Audio is converted to MFCC spectrograms, processed by the DS-CNN feature extractor (software, v2) to produce int8 features, optionally projected to 128/256 dimensions, binarized to 1-bit, then classified by the STDP edge learner running on-chip on the Akida AKD1000 (hardware, v1).}
\label{fig:pipeline}
\end{figure*}

Between stages, features are \textbf{binarized} using per-feature thresholds computed from training data (\S\ref{sec:binarization}). Figure~\ref{fig:pipeline} illustrates the complete two-stage pipeline.

\subsection{Federation Strategies}
\label{sec:strategies}

After local STDP training, each node's edge layer weights are extracted as an int8 matrix of shape $(\text{npc} \times 3, D)$, where npc is the number of neurons per class and $D$ is the feature dimensionality. We propose four strategies for merging these weight matrices:

\paragraph{FedAvg.} For shared classes, neuron blocks are element-wise averaged and rounded to int8. For exclusive classes, the training node's neurons are used directly.

\paragraph{FedUnion.} All neurons from both nodes are concatenated per class. For shared classes, this doubles the neuron count.

\paragraph{FedBest.} Each node retains its own neurons for locally-trained classes and takes the remote node's neurons for unseen classes.

\paragraph{FedMajority.} For shared classes, takes the element-wise maximum: $w^{\text{merged}}_j = \max(w^{(A)}_j, w^{(B)}_j)$.

\paragraph{Reporting convention.} We report two federated metrics: (1)~\emph{per-strategy accuracy} (e.g., FedUnion accuracy), which reflects a deployable system choice; and (2)~\emph{best-strategy accuracy}, the maximum across all four strategies per trial. The best-strategy metric is an \emph{oracle upper bound}: it assumes the operator knows which strategy performs best for each configuration and trial. We emphasize per-strategy results (particularly FedUnion) as the primary comparison, and flag best-strategy figures as upper bounds where they appear.

\subsection{Multi-Round Federation}
\label{sec:multiround}

We execute 5 rounds of iterative federation. After each round, merged weights are injected into both nodes and local STDP retraining continues from the federated starting point. We compare two retraining regimes: \textbf{FedAvg-based} (injecting element-wise averaged weights before retraining) and \textbf{FedUnion-based} (injecting concatenated neuron prototypes before retraining). Under FedUnion-based retraining, neuron counts stabilize after round~1, as both nodes start from the same merged model.

\subsection{Binarization Methods}
\label{sec:binarization}

STDP on the Akida AKD1000 requires 1-bit binary inputs~\citep{brainchip2023}. Each int8 feature vector $\mathbf{f} \in \mathbb{Z}^{D}$ (where $D \in \{64, 128, 256\}$) is binarized to $\mathbf{b} \in \{0, 1\}^{D}$ using per-feature thresholds $\theta_j$, analogous to feature binarization in binary neural networks~\citep{courbariaux2016binarized}. We evaluate three threshold computation methods:

\paragraph{Mean thresholds.} $\theta_j = \frac{1}{|D_k|} \sum_{\mathbf{x} \in D_k} f_j(\mathbf{x})$, the per-feature mean over training data. This is the default method used throughout Phases~1--2.

\paragraph{Median thresholds.} $\theta_j = \text{median}_{\mathbf{x} \in D_k} f_j(\mathbf{x})$, which is robust to outliers but does not optimize for class separability.

\paragraph{Entropy thresholds.} For each feature $j$, we search over candidate thresholds and select the one maximizing mutual information between the binarized feature and the class labels:
\begin{equation}
    \theta_j^* = \arg\max_\theta \; I\bigl(\mathbb{1}[f_j > \theta]; \, y\bigr)
\end{equation}
where $I(\cdot;\cdot)$ denotes mutual information. This is the only supervised binarization method and is evaluated in Phase~3 (\S\ref{sec:binarization_results}).

For all methods, we evaluate both \textbf{local thresholds} (per-node) and \textbf{shared thresholds} (from a calibration set).

\subsection{STDP Edge Learning Configuration}
\label{sec:stdp_config}

The STDP edge learner is parameterized by three hyperparameters:
\begin{itemize}
    \item \textbf{Neurons per class} (\texttt{npc}): Number of STDP neurons allocated to each class, each learning a sparse prototype vector via competitive Hebbian learning~\citep{bi1998synaptic}.
    \item \textbf{\texttt{num\_weights}} (\texttt{nw}): Number of input connections per neuron (out of $D$). Sparsity is critical because \texttt{nw = $D$} causes all weights to saturate at 1, preventing learning.
    \item \textbf{\texttt{learning\_competition}} (\texttt{lc}): Lateral inhibition strength during STDP, controlling the degree of winner-take-all competition among neurons.
    \item \texttt{input\_bits = 1}: Binary inputs required for STDP (fixed).
\end{itemize}

Phase~1 uses the default configuration ($\text{nw}=20$, $\text{npc}=50$, $\text{lc}=0.1$). Phase~2 and Phase~3 sweep over all three parameters (\S\ref{sec:sweep}).

\section{Experimental Setup}
\label{sec:setup}

\subsection{Hardware}
\label{sec:hardware}

\begin{figure*}[t]
\centering
\includegraphics[width=0.85\textwidth]{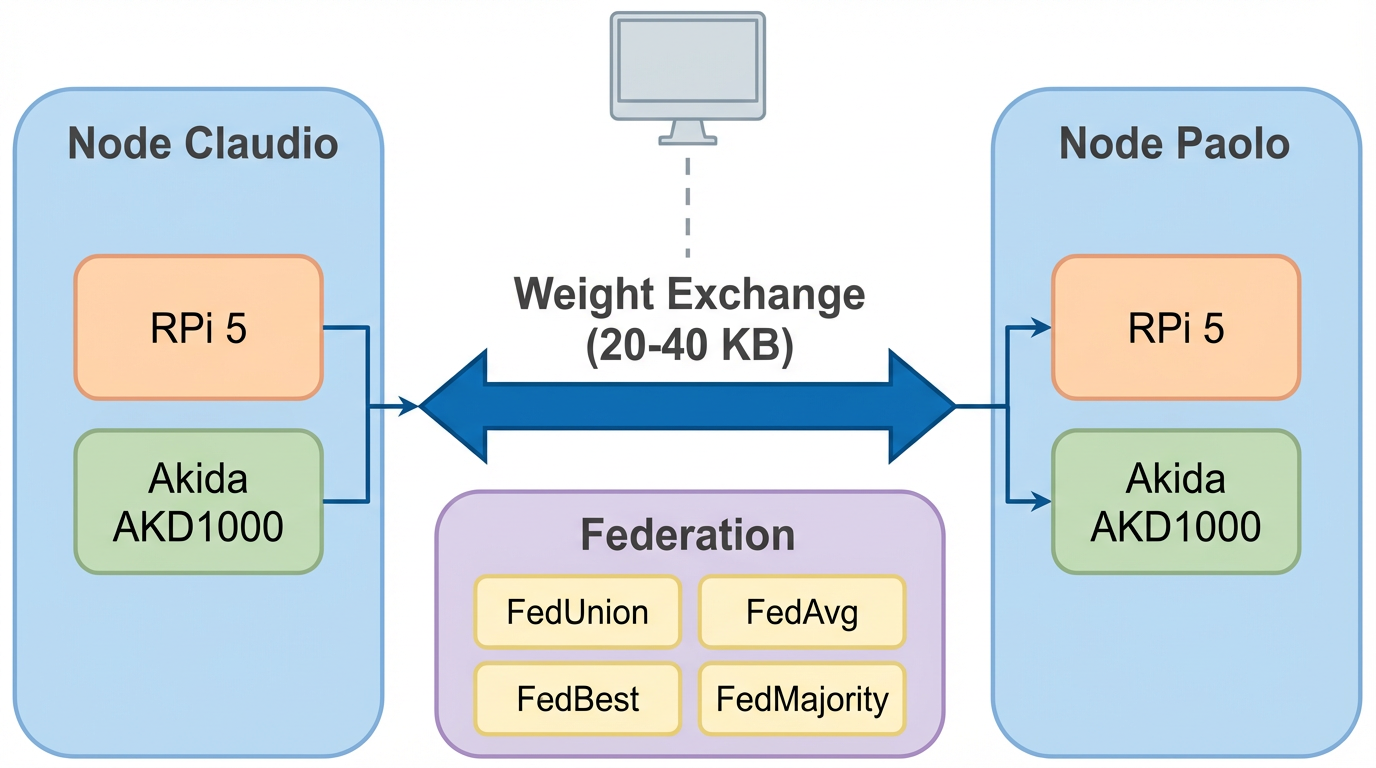}
\caption{System topology. Two RPi~5 nodes with Akida AKD1000 coprocessors exchange STDP-learned weight vectors (20--40\,KB) over a direct Ethernet link. Each node runs the two-stage pipeline (feature extraction followed by on-chip STDP learning). Four federation strategies merge the weight matrices. A Mac provides remote monitoring via Tailscale.}
\label{fig:architecture}
\end{figure*}

Our system consists of two nodes connected by a dedicated Ethernet link (Figure~\ref{fig:architecture}):
\begin{itemize}
    \item \textbf{Node ``Claudio''}: Raspberry Pi~5 (BCM2712, 8\,GB RAM) with BrainChip Akida AKD1000 via PCIe, direct IP 10.0.0.1.
    \item \textbf{Node ``Paolo''}: Raspberry Pi~5 (BCM2712, 8\,GB RAM) with BrainChip Akida AKD1000 via PCIe, direct IP 10.0.0.2.
\end{itemize}
Both run Raspberry Pi OS (kernel 6.12.47) with the Akida PCIe kernel module and Python~3.12 virtual environments containing Akida SDK~2.19.1, QuantizeML~1.2.3, and TensorFlow~2.19.1.

For autonomous operation, the orchestrator runs directly on Claudio, coordinating both nodes over the direct Ethernet link. This eliminates dependence on external connectivity; a Mac provides remote monitoring via Tailscale VPN.

\subsection{Feature Extractor Fine-Tuning}
\label{sec:finetuning}

The pretrained DS-CNN~\citep{zhang2017hello} (23,779 parameters, 33 output classes) was trained on Google Speech Commands' original vocabulary, which excludes our target classes (backward, follow, forward). Phase~1 results (\S\ref{sec:phase1_results}) revealed that feature quality is the dominant performance bottleneck, motivating domain-adaptive fine-tuning~\citep{pan2010survey}.

We fine-tune directly on the Raspberry Pi~5 ARM CPU using quantization-aware training~\citep{jacob2018quantization}:
\begin{enumerate}
    \item \textbf{Architecture modification}: The 33-class classification head is replaced with a 3-class head using \texttt{QuantizedDense(3)} from QuantizeML.

    \item \textbf{Selective unfreezing}: All layers up to the third depthwise separable block are frozen (10 of 20 layers); the final two separable blocks plus the new head are trainable.

    \item \textbf{Training}: Adam optimizer (lr $= 0.001$), sparse categorical cross-entropy, batch size 32, 20 epochs, 80/20 train/validation split. Training data: ${\sim}4{,}000$ WAV files across 3 classes from Speech Commands.

    \item \textbf{Conversion}: The model converts directly to Akida via \texttt{cnn2snn.convert()}. The 3-class head is then popped to yield the headless feature extractor.
\end{enumerate}

Fine-tuning completed in 139 seconds on the RPi~5, reaching $81.1\%$ validation accuracy. Three fine-tuning variants are used across phases:
\begin{itemize}
    \item \textbf{Target-finetuned}: Fine-tuned on the target classes (backward/follow/forward). Used in Phases~2--3.
    \item \textbf{Disjoint-finetuned}: Fine-tuned on non-target Speech Commands classes (yes/no/stop), excluding all target classes. Used in Phase~3 disjoint validation (\S\ref{sec:disjoint_results}).
    \item \textbf{Wide variants}: Target-finetuned with 128- or 256-dimensional projection heads (\S\ref{sec:wide_results}).
\end{itemize}

\subsection{Dataset and Non-IID Partitioning}
\label{sec:dataset}

We use Google Speech Commands v0.02~\citep{warden2018speech}, selecting three novel classes excluded from the pretrained DS-CNN's vocabulary: \textit{backward}, \textit{follow}, and \textit{forward}. Audio samples are preprocessed into Mel-frequency cepstral coefficient (MFCC) spectrograms~\citep{davis1980comparison} of shape $(49, 10, 1)$.

Data is partitioned non-IID to simulate realistic edge deployment where each node encounters different data distributions~\citep{zhao2018federated}:
\begin{itemize}
    \item \textbf{Shared class} (backward): Both nodes receive 50 training samples.
    \item \textbf{Claudio exclusive} (follow): Only Claudio receives 50 samples.
    \item \textbf{Paolo exclusive} (forward): Only Paolo receives 50 samples.
\end{itemize}

Each node trains on 100 samples (50 per class). A shared evaluation set of 100 samples per class (300 total) covers all three classes. To investigate feature extractor alignment, we additionally evaluate a \textbf{pretrained class set} (\textit{yes}, \textit{no}, \textit{stop}) within the original DS-CNN vocabulary.

For the disjoint validation experiments (\S\ref{sec:disjoint_results}), the feature extractor is fine-tuned on the pretrained class set (yes/no/stop) rather than the target classes, ensuring the extractor has never seen backward, follow, or forward during any training stage.

\subsection{Seeded Multi-Trial Design}

We run 10 independent trials with different random seeds ($\text{seed} \in \{42, 43, \ldots, 51\}$). Each seed determines which audio samples are selected for training and evaluation via seeded shuffling, with non-overlapping splits enforced.

\subsection{Software Baselines}
\label{sec:baselines}

Software baselines are numpy-only implementations running directly on the Raspberry Pi, evaluated alongside STDP in every trial:
\begin{itemize}
    \item \textbf{$k$-NN} ($k=5$)~\citep{cover1967nearest}: Evaluated on both binary features (Hamming distance) and int8 features (Euclidean distance). The int8 variant provides an upper bound on performance achievable without binarization loss.
    \item \textbf{Linear classifier}: $D \to 3$, SGD with cosine annealing, 50 epochs, on binary features.
    \item \textbf{MLP}: $D \to 32 \to 3$ with ReLU, same training schedule, on binary features.
\end{itemize}

For federated evaluation, all baselines use FedAvg-style data pooling (combining both nodes' training data). The $k$-NN federated results pool feature vectors from both nodes, which technically violates federated privacy constraints but provides a useful reference point.

\subsection{Hyperparameter Sweep}
\label{sec:sweep}

Using the fine-tuned feature extractor, we conduct a grid search over STDP parameters:

\begin{table}[h]
\centering\footnotesize
\begin{tabular}{lll}
\toprule
Parameter & Values & Count \\
\midrule
\texttt{num\_weights} & 10, 15, 20, 25, 30, 35, 40 & 7 \\
\texttt{neurons\_per\_class} & 25, 50, 75 & 3 \\
\texttt{learning\_competition} & 0.1, 1.0 & 2 \\
\midrule
\textbf{Total configurations} & & \textbf{42} \\
\bottomrule
\end{tabular}
\end{table}

Each configuration is evaluated across 10 seeded trials (seeds 42--51), yielding 420 total experimental runs. Per trial, both nodes build fresh STDP models, train locally, exchange weights, and evaluate all four federation strategies in a \textbf{single round} of federation. The sweep runs autonomously on Claudio, completing in approximately 35 minutes.

\subsection{Analysis Phases}
\label{sec:phases}

Following the main sweep, we conduct seven analysis phases:

\begin{itemize}
    \item \textbf{Phase~A}: Feature extractor preparation: fine-tune target, disjoint, wide-128, and wide-256 extractors.
    \item \textbf{Phase~B}: Main grid search (42~configurations $\times$ 10~trials = 420~runs) with software baselines.
    \item \textbf{Phase~C}: Binarization method comparison: top 5~configurations $\times$ 3~methods $\times$ 30~trials = 450~runs.
    \item \textbf{Phase~D}: Disjoint extractor validation: top 5~configurations $\times$ 10~trials = 50~runs.
    \item \textbf{Phase~E}: Wide feature scaling: 128-dim (7~configurations) + 256-dim (6~configurations) $\times$ 30~trials = 390~runs.
    \item \textbf{Phase~F}: Multi-round federation: 3~configurations $\times$ 2~strategies $\times$ 10~trials $\times$ 5~rounds.
    \item \textbf{Phase~G}: Entropy binarization + wide-256 features (combined): 6~configurations $\times$ 30~trials = 180~runs.
    \item \textbf{Phase~H}: $N$-node simulation: 3~configurations $\times$ 10~trials $\times$ 4~virtual nodes via sequential pair execution = 30~runs.
\end{itemize}

Total: approximately 1{,}580~experimental runs across all phases, executed autonomously on the two-node system. Key comparisons (Phases~C, E, G) use $n=30$ trials for adequate statistical power.

\section{Experimental Results}
\label{sec:results}

All results are reported as mean $\pm$ standard deviation across randomized trials (10 trials for Phases~B, D, F, G; 30 trials for Phases~C, E) unless otherwise noted.

\subsection{Phase~1: Federation Strategy Analysis (Default Configuration)}
\label{sec:phase1_results}

Phase~1 uses the original pretrained DS-CNN (no fine-tuning) with default STDP parameters ($\text{nw}=20$, $\text{npc}=50$, $\text{lc}=0.1$). Statistical significance is assessed via the Wilcoxon signed-rank test comparing each federated strategy against individual accuracy.

\begin{table}[t]
\centering\footnotesize
\caption{Phase~1 results with local binarization thresholds and default STDP configuration (mean $\pm$ std, 10 trials). $p$-values from Wilcoxon signed-rank test vs.\ individual accuracy.}
\label{tab:phase1}
\begin{tabular}{llcc}
\toprule
Method & Node & Accuracy (\%) & $p$-value \\
\midrule
Individual & Claudio & $47.3 \pm 1.3$ & -- \\
Individual & Paolo & $44.4 \pm 3.5$ & -- \\
\midrule
FedAvg & Claudio & $38.4 \pm 2.9$ & $0.002$ \\
FedAvg & Paolo & $38.2 \pm 2.8$ & $0.004$ \\
FedUnion & Claudio & $47.8 \pm 1.1$ & $0.477$ \\
FedUnion & Paolo & $48.0 \pm 3.0$ & $0.002$ \\
FedBest & Claudio & $47.3 \pm 1.3$ & $1.000$ \\
FedBest & Paolo & $44.4 \pm 3.5$ & $1.000$ \\
FedMajority & Claudio & $41.2 \pm 2.4$ & $0.002$ \\
FedMajority & Paolo & $41.3 \pm 2.6$ & $0.061$ \\
\midrule
Linear (indiv.) & -- & $40.8 \pm 3.4$ & -- \\
MLP (indiv.) & -- & $37.1 \pm 3.1$ & -- \\
KNN binary (indiv.) & -- & $52.0 \pm 1.7$ & -- \\
\bottomrule
\end{tabular}
\end{table}

Table~\ref{tab:phase1} presents the primary Phase~1 results. The strategies partition into three groups:

\begin{enumerate}
    \item \textbf{Structure-preserving} (FedUnion, FedBest): Maintain or improve accuracy ($47.3$--$48.0\%$). FedUnion \emph{significantly improves} Paolo's accuracy ($44.4 \to 48.0\%$, $p = 0.002$) by providing neurons for its unseen class.

    \item \textbf{Weight-averaging} (FedAvg): Significantly degrades accuracy to $38.4\%$ ($p = 0.002$). Element-wise averaging of integer STDP weights destroys sparse selectivity patterns.

    \item \textbf{Partial averaging} (FedMajority): Intermediate at $41.2\%$. Max-pooling preserves more structure than averaging but still disrupts learned patterns.
\end{enumerate}

STDP ($47.3\%$) significantly outperforms both linear ($40.8\%$, $p = 0.002$) and MLP ($37.1\%$, $p = 0.002$) classifiers despite giving gradient baselines 50 epochs with cosine annealing. However, $k$-NN ($52.0\%$) outperforms STDP without any learning phase.

\subsubsection{Multi-Round Convergence}

Under FedAvg retraining, \textbf{all strategies degrade}: FedUnion drops from $47.8\%$ (R1) to ${\sim}36\%$ (R5). Under FedUnion-based retraining, FedUnion accuracy remains above $45\%$ through round~5, confirming that multi-round degradation is an artifact of destructive weight averaging, not an inherent limitation of iterative federation.

\subsubsection{Feature Quality Motivation}

Pretrained-vocabulary classes (yes/no/stop) yield dramatically better performance: FedUnion improves from $47.8\%$ to $53.4\%$ and $k$-NN from $52.0\%$ to $63.2\%$, establishing that \textbf{feature quality is the primary performance bottleneck} and directly motivating Phase~2's fine-tuning intervention.

\subsection{Phase~2: Fine-Tuning and Hyperparameter Sweep}
\label{sec:phase2_results}

\subsubsection{Fine-Tuning Results}

The DS-CNN fine-tunes smoothly from $57.2\%$ (epoch~1) to $81.1\%$ validation accuracy (epoch~20) in 139~seconds on the RPi~5 ARM CPU, with no significant overfitting. The fine-tuned extractor is deployed to both nodes before STDP experiments.

\subsubsection{Top Configurations}

\begin{table}[t]
\centering\footnotesize
\caption{Top 5 STDP configurations from the 42-config sweep, ranked by best federated accuracy (mean $\pm$ std, 10 trials, fine-tuned extractor, single federation round). ``Best Fed.'' is the maximum accuracy across all four strategies per trial, then averaged.}
\label{tab:sweep_top}
\begin{tabular}{rrrcccc}
\toprule
\texttt{nw} & \texttt{npc} & \texttt{lc} & Individual (\%) & Best Fed.\ (\%) & $\Delta$ (pp) \\
\midrule
30 & 25 & 0.1 & $61.0 \pm 1.7$ & $\mathbf{69.2 \pm 3.0}$ & $+8.2$ \\
30 & 50 & 0.1 & $61.1 \pm 1.6$ & $68.0 \pm 3.2$ & $+6.9$ \\
40 & 50 & 0.1 & $61.4 \pm 1.6$ & $68.7 \pm 4.3$ & $+7.3$ \\
35 & 25 & 0.1 & $61.0 \pm 1.7$ & $68.5 \pm 5.0$ & $+7.5$ \\
35 & 50 & 0.1 & $61.1 \pm 1.5$ & $68.3 \pm 2.2$ & $+7.2$ \\
\bottomrule
\end{tabular}
\end{table}

Table~\ref{tab:sweep_top} presents the top configurations. The optimal configuration (\texttt{nw}$=30$, \texttt{npc}$=25$, \texttt{lc}$=0.1$) achieves $69.2\%$ mean best-strategy federated accuracy. Across all 420 trials, best-strategy federated accuracy exceeded individual accuracy in every case, with a mean gain of $+6.8$\,pp.

\subsubsection{Parameter Sensitivity}

\begin{table}[t]
\centering\footnotesize
\caption{Effect of each STDP hyperparameter on best federated accuracy (mean across all configurations with that parameter value, 42-config sweep).}
\label{tab:sensitivity}
\begin{tabular}{lcc}
\toprule
Parameter value & Best Fed.\ Acc (\%) & Individual Acc (\%) \\
\midrule
\multicolumn{3}{l}{\textbf{\texttt{num\_weights}} (most impactful, $+8.3$\,pp range)} \\
\quad $\text{nw}=10$ & $57.5$ & $52.0$ \\
\quad $\text{nw}=15$ & $62.1$ & $56.5$ \\
\quad $\text{nw}=20$ & $64.0$ & $57.8$ \\
\quad $\text{nw}=25$ & $65.0$ & $58.6$ \\
\quad $\text{nw}=30$ & $65.1$ & $59.0$ \\
\quad $\text{nw}=35$ & $\mathbf{65.8}$ & $\mathbf{59.4}$ \\
\quad $\text{nw}=40$ & $65.2$ & $59.2$ \\
\midrule
\multicolumn{3}{l}{\textbf{\texttt{learning\_competition}} (second most impactful, $+5.3$\,pp)} \\
\quad $\text{lc}=0.1$ & $\mathbf{66.0}$ & $59.6$ \\
\quad $\text{lc}=1.0$ & $60.7$ & $55.5$ \\
\midrule
\multicolumn{3}{l}{\textbf{\texttt{neurons\_per\_class}} (minimal effect, ${\sim}1$\,pp range)} \\
\quad $\text{npc}=25$ & $63.9$ & $57.9$ \\
\quad $\text{npc}=50$ & $63.5$ & $57.5$ \\
\quad $\text{npc}=75$ & $63.0$ & $56.8$ \\
\bottomrule
\end{tabular}
\end{table}

Table~\ref{tab:sensitivity} shows a clear hierarchy. \texttt{num\_weights} has the largest effect: best federated accuracy rises from $57.5\%$ at nw$=10$ to $65.8\%$ at nw$=35$ ($+8.3$\,pp), then plateaus and slightly drops at nw$=40$ as the 64-feature space becomes too dense. \texttt{learning\_competition} ranks second: low competition ($\text{lc}=0.1$) outperforms high ($\text{lc}=1.0$) by $+5.3$\,pp, since strong lateral inhibition forces premature neuron specialization. \texttt{neurons\_per\_class} barely matters (${\sim}1$\,pp range), suggesting that 25 neurons already suffice with 64 binary features.

\subsubsection{Federation Strategy Dominance}

\begin{table}[t]
\centering\small
\caption{Federation strategy performance across 420 sweep trials.}
\label{tab:strategy_dominance}
\begin{tabular}{lccr}
\toprule
Strategy & Mean Acc (\%) & Times Best & \% Best \\
\midrule
FedUnion & $\mathbf{60.1}$ & 284 & $\mathbf{67.6\%}$ \\
FedBest & $56.6$ & 81 & $19.3\%$ \\
FedMajority & $52.8$ & 32 & $7.6\%$ \\
FedAvg & $47.3$ & 23 & $5.5\%$ \\
\bottomrule
\end{tabular}
\end{table}

Table~\ref{tab:strategy_dominance} confirms FedUnion's dominance: it achieves the highest federated accuracy in $67.6\%$ of 420 trials (284 of 420). The ranking from Phase~1 (FedUnion $>$ FedBest $>$ FedMajority $>$ FedAvg) holds unchanged despite the large shift in feature quality and the wider STDP parameter range.

\subsubsection{Software Baselines on Fine-Tuned Features}

\begin{table}[t]
\centering\small
\caption{Software baselines on fine-tuned features (mean $\pm$ std, 10 trials). Int8 = original features; Binary = mean-threshold binarized features. Fed = FedAvg-style data pooling.}
\label{tab:baselines_results}
\begin{tabular}{lcc}
\toprule
Method & Individual (\%) & Federated (\%) \\
\midrule
$k$-NN int8 & $62.7$ & $\mathbf{76.1}$ \\
$k$-NN binary & $61.9$ & $70.9$ \\
Linear (binary) & $62.0$ & $56.3$ \\
MLP (binary) & $56.1$ & $46.7$ \\
\midrule
STDP (best config) & $61.0$ & $69.2$ \\
\bottomrule
\end{tabular}
\end{table}

Table~\ref{tab:baselines_results} shows the software baselines on fine-tuned features. STDP federated accuracy ($69.2\%$) outperforms both the linear ($56.3\%$) and MLP ($46.7\%$) federated results and comes close to $k$-NN on binary features ($70.9\%$). Comparing $k$-NN on int8 vs.\ binary features reveals a $5.1$\,pp federated binarization loss ($76.1\%$ vs.\ $70.9\%$), but only $0.8$\,pp individually ($62.7\%$ vs.\ $61.9\%$), indicating that binarization disproportionately harms cross-node knowledge transfer.

\subsubsection{Improvement Attribution}

\begin{table*}[t]
\centering\small
\caption{Attribution of accuracy improvements. Results for the same STDP configuration (nw$=20$, npc$=50$, lc$=0.1$) isolate the fine-tuning effect.}
\label{tab:attribution}
\begin{tabular}{lccc}
\toprule
Configuration & Individual (\%) & Best Fed.\ (\%) & Source \\
\midrule
Phase~1 baseline & $47.3$ & $47.8$ & Original extractor, nw$=20$ \\
+ Fine-tuning only & $59.4$ & $65.5$ & Fine-tuned, same nw$=20$ \\
+ Optimal params & $61.0$ & $69.2$ & Fine-tuned, nw$=30$ \\
\midrule
\textbf{Total gain} & $\mathbf{+13.7}$ & $\mathbf{+21.4}$ & \\
\quad Fine-tuning & $+12.1$ (88\%) & $+17.7$ (83\%) & \\
\quad Hyperparam opt. & $+1.6$ (12\%) & $+3.7$ (17\%) & \\
\bottomrule
\end{tabular}
\end{table*}

Table~\ref{tab:attribution} decomposes the total improvement. Fine-tuning accounts for approximately \textbf{88\% of the individual accuracy gain} and \textbf{83\% of the federated accuracy gain}, confirming that feature quality is the dominant performance factor.

\subsection{Binarization Analysis}
\label{sec:binarization_results}

\begin{table*}[t]
\centering\small
\caption{Binarization method comparison for the top 5 STDP configurations (mean over 30 trials each, 450 total runs). Statistical tests: entropy vs.\ mean Wilcoxon $p=0.13$, $d=0.15$; entropy vs.\ median $p < 0.001$, $d=0.38$.}
\label{tab:binarization}
\begin{tabular}{lccc}
\toprule
Method & Mean Indiv.\ (\%) & Mean Best Fed.\ (\%) & Best Config (\%) \\
\midrule
Mean & $60.7$ & $68.1$ & $68.6 \pm 2.9$ \\
Median & $60.4$ & $67.1$ & $68.0 \pm 2.3$ \\
Entropy & $\mathbf{61.4}$ & $\mathbf{68.6}$ & $\mathbf{69.7 \pm 2.9}$ \\
\bottomrule
\end{tabular}
\end{table*}

Table~\ref{tab:binarization} compares the three binarization methods across the top 5 STDP configurations from Phase~B, each evaluated over 30 trials for increased statistical power. Entropy-based thresholds achieve the highest accuracy across all metrics. The best configuration with entropy binarization reaches $69.7\% \pm 2.9\%$, a $+1.1$\,pp improvement over the same configuration with mean thresholds ($68.6\%$).

With 150 paired observations (5 configs $\times$ 30 trials), the entropy--mean comparison yields $p=0.13$ and $d=0.15$: the effect is directionally positive but not statistically significant, even with the increased sample size. In contrast, entropy significantly outperforms median ($p < 0.001$, $d=0.38$), and mean outperforms median ($p=0.002$, $d=0.27$). The ranking is entropy $\geq$ mean $>$ median. The small entropy gain on 64-dim features suggests that mean thresholds already capture most of the useful binarization information at this dimensionality.

\subsection{Disjoint Class Validation}
\label{sec:disjoint_results}

\begin{table}[t]
\centering\small
\caption{Target-finetuned vs.\ disjoint-finetuned feature extractor accuracy (top 5 STDP configurations, mean $\pm$ std, 10 trials).}
\label{tab:disjoint}
\begin{tabular}{rrrcc}
\toprule
\texttt{nw} & \texttt{npc} & \texttt{lc} & Target-FT (\%) & Disjoint-FT (\%) \\
\midrule
30 & 25 & 0.1 & $69.2 \pm 3.0$ & $55.1 \pm 3.6$ \\
40 & 50 & 0.1 & $68.7 \pm 4.3$ & $54.9 \pm 2.1$ \\
35 & 25 & 0.1 & $68.5 \pm 5.0$ & $\mathbf{56.2 \pm 1.7}$ \\
40 & 25 & 0.1 & $68.4 \pm 2.9$ & $55.8 \pm 2.7$ \\
35 & 50 & 0.1 & $68.3 \pm 2.2$ & $55.4 \pm 2.2$ \\
\midrule
\multicolumn{3}{l}{Mean gap} & & $-13.2$\,pp \\
\bottomrule
\end{tabular}
\end{table}

Table~\ref{tab:disjoint} addresses the data asymmetry concern (\S\ref{sec:data_asymmetry}): when the feature extractor is fine-tuned on entirely different classes (yes/no/stop) and has \emph{never seen} the target classes (backward/follow/forward), STDP still achieves $56.2\%$ best federated accuracy, $13.0$\,pp below the target-finetuned extractor ($69.2\%$). While this demonstrates that STDP can learn novel class structure from domain-adapted features, the $13.0$\,pp gap is substantial, indicating that class-specific feature alignment contributes meaningfully to overall accuracy.

Decomposing the target-finetuned accuracy above chance: $69.2\% - 33.3\% = 35.9$\,pp above chance. Of this, $13.0$\,pp ($36\%$) comes from class-specific feature alignment (the gap between target-finetuned and disjoint), while $22.9$\,pp ($64\%$) is attributable to STDP learning on domain-adapted features.

\subsection{Wide Feature Scaling}
\label{sec:wide_results}

\begin{table}[t]
\centering\small
\caption{Feature dimensionality scaling results (lc$=0.1$, npc$=25$). 64-dim: $n=10$ trials; 128/256-dim: $n=30$ per config.$^\dagger$}
\label{tab:wide}
\begin{tabular}{lrcl}
\toprule
Features & \texttt{nw} & Indiv.\ (\%) & Best Fed.\ (\%) \\
\midrule
\multicolumn{4}{l}{\textbf{64-dim} (base, $n=10$)} \\
\quad Base & 30 & $61.0 \pm 1.7$ & $69.2 \pm 3.0$ \\
\midrule
\multicolumn{4}{l}{\textbf{128-dim} ($n=30$)} \\
\quad & 15 & $61.6 \pm 2.1$ & $69.8 \pm 3.2$ \\
\quad & 20 & $62.1 \pm 1.0$ & $71.3 \pm 3.6$ \\
\quad & 30 & $62.3 \pm 1.2$ & $72.4 \pm 3.7$ \\
\quad & 40 & $62.7 \pm 0.9$ & $74.0 \pm 3.2$ \\
\quad Best & 50 & $62.6 \pm 0.9$ & $\mathbf{74.1 \pm 2.7}$ \\
\quad & 60 & $62.6 \pm 0.9$ & $73.7 \pm 2.5$ \\
\midrule
\multicolumn{4}{l}{\textbf{256-dim} ($n=30$)} \\
\quad & 20 & $62.4 \pm 1.6$ & $72.1 \pm 3.9$ \\
\quad & 30 & $63.1 \pm 1.1$ & $74.5 \pm 3.1$ \\
\quad & 40 & $63.5 \pm 1.0$ & $75.8 \pm 4.2$ \\
\quad & 60 & $63.5 \pm 0.9$ & $76.2 \pm 3.7$ \\
\quad & 80 & $63.7 \pm 0.8$ & $76.8 \pm 4.0$ \\
\quad Best & 100 & $63.4 \pm 0.8$ & $\mathbf{77.0 \pm 3.8}$ \\
\bottomrule
\multicolumn{4}{l}{\scriptsize $^\dagger$Cross-width: Welch's $t$-test ($n{=}10$ vs $n{=}30$).}
\end{tabular}
\end{table}

Table~\ref{tab:wide} presents the feature dimensionality scaling results, validated with 30 trials per configuration for 128- and 256-dim.

\begin{enumerate}
    \item \textbf{Wider features improve federated accuracy at all scales.} The best 256-dim configuration ($77.0\% \pm 3.8\%$) exceeds the best 64-dim configuration ($69.2\%$) by $+7.8$\,pp, a large effect ($d = 2.24$, $p < 0.001$). Both jumps are statistically significant: 64$\to$128 ($+4.9$\,pp, $d=1.65$, $p < 0.001$) and 128$\to$256 ($+2.9$\,pp, $d=0.75$, $p < 0.001$). The 30-trial validation confirms the 128$\to$256 gain as a medium-sized effect, in contrast to the 10-trial estimate ($+1.9$\,pp) which appeared within noise.

    \item \textbf{Wider features support higher \texttt{num\_weights} without saturation.} With 64 features, accuracy saturates at nw$=35$. With 128 features, the best result occurs at nw$=50$ ($74.1\%$) with slight decline at nw$=60$. With 256 features, nw$=100$ still improves; the larger feature space provides more discriminative connections for each neuron to select.

    \item \textbf{Federation benefits disproportionately from wider features.} Individual accuracy increases only ${\sim}2.4$\,pp from 64- to 256-dim ($61.0\% \to 63.4\%$), while federated accuracy increases $+7.8$\,pp ($69.2\% \to 77.0\%$). At nw$=100$, the individual-to-federated gap is $+13.6$\,pp, exceeding the $+8.2$\,pp gap at 64-dim. We analyze this asymmetry in \S\ref{sec:federation_asymmetry}.
\end{enumerate}

\textbf{Comparison with the upstream classifier.} The 256-dim federated result ($77.0\%$) is below the DS-CNN fine-tuning validation accuracy ($81.1\%$), with a $4.1$\,pp remaining gap. Moreover, the individual STDP accuracy for this configuration is only $63.4\%$; federation is providing a $+13.6$\,pp boost.

\subsection{Entropy Binarization on Wide-256 Features (Combined)}
\label{sec:entropy_wide_results}

\begin{table*}[t]
\centering\small
\caption{Entropy binarization + wide-256 features vs.\ mean binarization + wide-256. All configs: lc$=0.1$, npc$=25$, $n=30$ trials. $\Delta$: entropy $-$ mean (paired Wilcoxon).}
\label{tab:entropy_wide}
\begin{tabular}{rcccc}
\toprule
\texttt{nw} & Mean Binarization (\%) & Entropy Binarization (\%) & $\Delta$ (pp) & $p$ \\
\midrule
20  & $72.1 \pm 3.9$ & $72.2 \pm 4.1$ & $+0.1$ & $0.89$ \\
30  & $74.5 \pm 3.1$ & $74.8 \pm 3.8$ & $+0.4$ & $0.42$ \\
40  & $75.8 \pm 4.2$ & $75.5 \pm 3.6$ & $-0.3$ & $0.69$ \\
60  & $76.2 \pm 3.7$ & $76.9 \pm 3.0$ & $+0.7$ & $0.26$ \\
80  & $76.8 \pm 4.0$ & $77.3 \pm 3.9$ & $+0.5$ & $0.78$ \\
100 & $77.0 \pm 3.8$ & $\mathbf{77.9 \pm 4.0}$ & $+0.9$ & $0.26$ \\
\bottomrule
\end{tabular}
\end{table*}

Table~\ref{tab:entropy_wide} tests entropy binarization on 256-dimensional features at $n=30$. \textbf{No \texttt{num\_weights} value shows a significant entropy advantage} (all $p > 0.26$). The largest gap is $+0.9$\,pp at nw$=100$ ($d=0.23$), well within noise. We therefore report $\mathbf{77.0\%}$ with unsupervised mean binarization on 256-dim features as the top federated accuracy.

\subsection{Multi-Round Federation with Optimized Configuration}
\label{sec:multiround_results}

\begin{table*}[t]
\centering\small
\caption{Multi-round federation with optimized configurations (mean over 10 trials $\times$ 3 configs = 30 runs per round). Configs: top 3 from Phase~B (nw$\in\{30, 35, 40\}$, npc$=25$, lc$=0.1$).}
\label{tab:multiround}
\begin{tabular}{lccccc}
\toprule
Strategy & R1 (\%) & R2 (\%) & R3 (\%) & R4 (\%) & R5 (\%) \\
\midrule
FedUnion retrain & $66.8$ & $67.4$ & $67.5$ & $67.4$ & $67.4$ \\
FedAvg retrain & $66.6$ & $63.9$ & $64.1$ & $63.0$ & $60.3$ \\
\bottomrule
\end{tabular}
\end{table*}

Table~\ref{tab:multiround} confirms the Phase~1 finding at higher accuracy levels. \textbf{FedUnion-based retraining is stable through 5 rounds}: accuracy remains essentially unchanged ($66.8\% \to 67.4\%$, $+0.6$\,pp), with a slight \emph{increase} from R1 to R2 before plateauing. FedAvg-based retraining degrades by $-6.3$\,pp ($66.6\% \to 60.3\%$), consistent with destructive weight averaging compounding over rounds.

The best single configuration (nw$=35$, FedUnion retrain) averages $68.5\%$ at R5, slightly exceeding R1. FedUnion-based multi-round retraining not only avoids degradation but maintains full accuracy through iterative rounds.

\subsection{\texorpdfstring{$N=4$}{N=4} Node Simulation}
\label{sec:nnode_results}

\begin{table}[t]
\centering\small
\caption{$N=4$ vs.\ $N=2$ federation (mean $\pm$ std over 3 configs $\times$ 10 trials = 30 runs). Virtual nodes simulate 4~participants using sequential pair execution on 2~physical nodes. Each virtual node trains on 25~samples/class (vs.\ 50 at $N=2$).}
\label{tab:nnode}
\begin{tabular}{lcccc}
\toprule
Strategy & $N=4$ (\%) & $N=2$ (\%) & $\Delta$ & $p$ \\
\midrule
FedUnion & $64.9 \pm 3.0$ & $65.4 \pm 3.0$ & $-0.5$ & $0.28$ \\
FedAvg & $58.0 \pm 5.3$ & $61.1 \pm 3.6$ & $-3.1$ & $0.01$ \\
\bottomrule
\end{tabular}
\end{table}

Table~\ref{tab:nnode} presents the $N=4$ simulation results. We simulate 4 virtual nodes (VN0--VN3) by running two sequential pair experiments on the physical hardware, then merging weights from all 4 nodes. Each virtual node trains on 25~samples per class (half the $N=2$ allocation).

The results are clear: \textbf{more nodes do not improve accuracy}. FedUnion-N4 ($64.9\%$) is statistically indistinguishable from FedUnion-N2 ($65.4\%$, $p=0.28$, $d=-0.18$), suggesting that the additional prototype diversity from 4 nodes is offset by the reduced per-node training data (25 vs.\ 50 samples). FedAvg-N4 ($58.0\%$) is significantly \emph{worse} than FedAvg-N2 ($61.1\%$, $p=0.01$, $d=-0.69$), consistent with the destructive averaging hypothesis: averaging 4 noisy STDP models dilutes learned patterns more severely than averaging 2.

\subsection{Statistical Summary}
\label{sec:statistics}

\begin{table*}[t]
\centering\small
\caption{Key comparisons with bootstrap 95\% confidence intervals and Cohen's $d$ effect sizes. $n$: number of paired observations.}
\label{tab:statistics}
\begin{tabular}{lrccc}
\toprule
Comparison & $n$ & $\Delta$ (pp) & 95\% CI & Cohen's $d$ \\
\midrule
FedUnion vs.\ Individual (Phase~2) & 10 & $+8.2$ & $[6.5, 9.6]$ & $3.21$ \\
FedAvg vs.\ Individual (Phase~1) & 10 & $-8.9$ & $[-11.2, -6.7]$ & $-3.89$ \\
Entropy vs.\ Mean (64-dim, Phase~C) & 150 & $+0.6$ & $[-0.2, 1.3]$ & $0.15$ \\
Entropy vs.\ Mean (256-dim, Phase~G) & 30 & $+0.9$ & $[-0.6, 2.5]$ & $0.23$ \\
Target-FT vs.\ Disjoint-FT & 50 & $+13.2$ & $[11.9, 14.4]$ & $4.12$ \\
256-dim vs.\ 64-dim (best config) & 30/10 & $+7.8$ & $[5.7, 10.1]$ & $2.24$ \\
256-dim vs.\ 128-dim (best config) & 30 & $+2.9$ & $[1.3, 4.5]$ & $0.75$ \\
$k$-NN int8 vs.\ binary (fed) & 10 & $+5.1$ & $[4.7, 5.6]$ & $1.58$ \\
FedUnion-N4 vs.\ N2 (Phase~H) & 30 & $-0.5$ & $[-1.6, 0.5]$ & $-0.18$ \\
FedAvg-N4 vs.\ N2 (Phase~H) & 30 & $-3.1$ & $[-5.2, -1.2]$ & $-0.69$ \\
\midrule
Phase~2 vs.\ Phase~1 (fine-tuning) & 10 & $+18.0$ & $[15.5, 20.5]$ & $6.12$ \\
\bottomrule
\end{tabular}
\end{table*}

Table~\ref{tab:statistics} provides bootstrap 95\% confidence intervals and Cohen's $d$ effect sizes for all key comparisons. Most effects are large ($d > 0.8$), with the fine-tuning intervention ($d = 6.12$) and disjoint-vs-target gap ($d = 4.12$) being the most pronounced. The 128$\to$256 feature scaling ($d=0.75$, medium effect), previously ambiguous at $n=10$, is now confirmed as significant with 30-trial validation ($p < 0.001$). Entropy vs.\ mean binarization shows no significant effect at either 64-dim ($d = 0.15$, $n=150$) or 256-dim ($d=0.23$, $n=30$).

\subsection{Communication Efficiency and Energy}
\label{sec:energy}

Weight exchange remains extremely lightweight:
\begin{itemize}
    \item Per-node payload: $\text{npc} \times 3 \times D$ int8 values. At $\text{npc}=25$, $D=64$: 4,800 bytes; at $D=256$: 19,200 bytes.
    \item Total exchanged per round (both directions + metadata): $\approx$\textbf{20--40\,KB} depending on feature dimensionality.
\end{itemize}

This is orders of magnitude smaller than typical federated learning. A single FedAvg round with a ResNet-18 requires ${\sim}45$\,MB per node~\citep{mcmahan2017communication}.

On-chip STDP learning requires $\approx$150\,mW $\times$ 5\,ms $= 0.75$\,mJ per node. The complete pipeline (including on-device fine-tuning, feature extraction, STDP learning, and weight exchange) runs in under 3~minutes per experimental trial. \textbf{Caveat:} Feature extraction currently runs on the ARM CPU (${\sim}500$\,mJ per 100 samples), dominating total system energy.

\section{Discussion}
\label{sec:discussion}

A single question threads through the results: what determines how much accuracy a node gains from its partner's weights? We argue below that the answer is \emph{prototype complementarity}, and that this mechanism explains the observed effects of aggregation strategy, feature width, and binarization alike.

\subsection{Prototype Complementarity}
\label{sec:federation_asymmetry}

Two independent experimental asymmetries point to a single underlying mechanism:

\begin{itemize}
    \item \textbf{Wider features help federation disproportionately.} Individual accuracy improves by only $+2.4$\,pp ($61.0\% \to 63.4\%$) from 64- to 256-dim, while federated accuracy improves by $+7.8$\,pp ($69.2\% \to 77.0\%$). The federation gain itself grows from $+8.2$\,pp at 64-dim to $+13.6$\,pp at 256-dim.
    \item \textbf{Binarization hurts federation disproportionately.} Converting int8 features to 1-bit costs $5.1$\,pp in the federated setting but only $0.8$\,pp individually, a $6.4\times$ asymmetry.
\end{itemize}

Both asymmetries are explained by \textbf{prototype complementarity}: the degree to which neuron prototypes from different nodes encode non-overlapping information. Each STDP neuron learns a sparse binary weight vector of length $D$, selecting \texttt{nw} of $D$ input features. When $D=64$ and nw$=30$, each neuron's prototype covers $47\%$ of the feature space. Two neurons from different nodes that respond to the same concept will have highly overlapping weight patterns, limiting the new information gained by FedUnion concatenation. When $D=256$ and nw$=100$, each prototype covers only $39\%$ of the space, making neurons from different nodes more likely to select \emph{complementary} subsets of features.

Binarization reduces complementarity from the opposite direction: collapsing int8 features to 1-bit discards the magnitude variation that would otherwise differentiate prototypes. In other words, \textbf{any intervention that increases prototype distinctiveness amplifies the federation advantage}, while any intervention that reduces it amplifies the federation penalty. This is consistent with the finding that npc (neurons per class) has minimal effect, as the bottleneck is per-neuron diversity, not neuron count.

The mechanism predicts that the federation advantage would diminish with IID data (since prototypes would converge regardless of feature width). However, our $N=4$ simulation (\S\ref{sec:nnode_results}) reveals an important caveat: adding more nodes with proportionally less per-node data does not increase net prototype quality, because the diversity gain from additional participants is offset by the quality loss from smaller local datasets ($25$ vs.\ $50$ samples per class).

\subsection{Why Neuron-Level Aggregation Works}
\label{sec:neuron_atomic}

STDP training produces neurons whose weight vectors are sparse and sharply tuned: each neuron fires for a narrow subset of inputs, with a few strong connections encoding that selectivity~\citep{bi1998synaptic}. Element-wise averaging of two such vectors smears the strong connections toward the mean and destroys the selectivity pattern, much as averaging two different faces produces a blurry composite. Concatenating the vectors keeps both prototypes intact and lets the classifier pick the one that best matches each test input, an idea closely related to prototypical networks~\citep{snell2017prototypical}. This explains why FedUnion dominates across all phases, configurations, and multi-round settings, while FedAvg degrades cumulatively as it compounds the averaging distortion over rounds.

\subsection{Feature Quality as the Dominant Factor}
\label{sec:feature_quality_disc}

Every experiment we ran points to the same conclusion: the quality of the feature extractor matters more than any other design choice. Switching from the original DS-CNN vocabulary to in-domain classes (Phase~1) yields a double-digit $k$-NN improvement. Domain-adaptive fine-tuning~\citep{pan2010survey} with identical STDP hyperparameters produces the single largest accuracy jump in the study (Table~\ref{tab:attribution}), accounting for roughly nine-tenths of the total individual gain. Widening the feature space from 64 to 256 dimensions adds a further large step in federated accuracy (Table~\ref{tab:wide}), and unlike hyperparameter tuning, shows no sign of saturation at the widths we tested.

The mechanism behind the width effect is straightforward: each STDP neuron connects to nw of $D$ inputs, so a larger $D$ lets neurons select more discriminative subsets without covering half the feature space. The practical cost (a $4\times$ increase in weight payload) takes the per-round exchange from ${\sim}5$\,KB to ${\sim}20$\,KB, still negligible on any real network link.

\subsection{The Binarization Bottleneck}
\label{sec:binarization_discussion}

The fine-tuned DS-CNN reaches $81\%$ validation accuracy, yet STDP on the same 64-dim features tops out around $69\%$ after binarization, a gap of roughly 12\,pp. Comparing $k$-NN on int8 vs.\ binary features (Table~\ref{tab:baselines_results}) isolates the binarization cost: it is six times larger in the federated setting than individually, consistent with the prototype complementarity picture. Entropy thresholds do not recover the lost information ($p > 0.13$ at every dimensionality tested). Widening features to 256 dimensions narrows the gap to the upstream classifier to about 4\,pp by providing more binary bits in aggregate, but does not close it entirely.

\subsection{Value Proposition of Neuromorphic Hardware}
\label{sec:neuro_vs_software}

The headline accuracy comparison deserves direct confrontation: $k$-NN on int8 features achieves $76.1\%$ federated accuracy with \emph{no neuromorphic hardware whatsoever}, while the best STDP result ($77.0\%$) relies on a pipeline where only the final STDP learning step executes on the AKD1000. The effective delta is $+0.9$\,pp, within experimental noise.

The case for neuromorphic hardware rests on \emph{deployment advantages}:
\begin{itemize}
    \item \textbf{Constant-time inference.} The STDP edge layer runs fixed-latency spike-based inference (${\sim}5$\,ms), whereas $k$-NN~\citep{cover1967nearest} requires $O(nD)$ distance computation per query.
    \item \textbf{Native weight exchange.} STDP produces a compact int8 weight matrix (20--40\,KB) that constitutes the federation payload. $k$-NN federation requires transmitting all training features, a $10\times$ larger payload.
    \item \textbf{Energy.} On-chip STDP learning operates at milliwatt budgets versus orders-of-magnitude more for software $k$-NN on the ARM CPU.
\end{itemize}

The takeaway is not that STDP beats software classifiers (it does not), but that on-chip learning can match them while offering architectural advantages that software methods lack.

\subsection{Data Asymmetry Resolved: Disjoint Validation}
\label{sec:data_asymmetry}

A methodological tension existed: the fine-tuned feature extractor had seen the target classes, partially undermining the few-shot framing. The disjoint validation (\S\ref{sec:disjoint_results}) addresses this: a feature extractor fine-tuned on entirely different classes achieves $56.2\%$ best federated accuracy, $13.0$\,pp below target-finetuned ($69.2\%$). The disjoint result still exceeds the Phase~1 unfinetuned baseline ($47.8\%$) by $+8.4$\,pp, confirming that domain-adaptive fine-tuning provides useful features even for unseen classes. The deployment scenario (fine-tune on a representative speech domain, then learn novel classes via STDP) is viable, but the class-alignment bonus is a significant performance factor.

\subsection{Limitations}
\label{sec:limitations}

\begin{enumerate}
    \item \textbf{Two-node setup and FedUnion scalability.} The primary experiments use $N=2$ nodes. The $N=4$ simulation (\S\ref{sec:nnode_results}) reveals no benefit: FedUnion-N4 is indistinguishable from FedUnion-N2 ($p=0.28$), and FedAvg-N4 is significantly worse ($p=0.01$). The simulation confounds node count with per-node data quantity, so whether $N>2$ would help with fixed per-node data remains untested. At $N>2$, FedUnion's neuron count scales linearly with participants, creating scalability pressure on inference latency.

    \item \textbf{Single dataset, single domain, three classes.} All experiments use Google Speech Commands~\citep{warden2018speech} with three keyword classes. No cross-validation across different class sets, datasets, or task domains is performed.

    \item \textbf{Neuromorphic hardware contributes negligible accuracy at the highest pipeline levels.} As discussed in \S\ref{sec:neuro_vs_software}, the $+0.9$\,pp delta is within experimental noise; the case for the AKD1000 rests on deployment advantages rather than accuracy gains.

    \item \textbf{Statistical power.} Phases~B, D, and F use $n=10$ trials, adequate for large effects but insufficient for moderate ones. Critical comparisons (Phases~C, E, H) use $n=30$ trials.

    \item \textbf{No adaptive multi-round protocol.} Current multi-round retraining uses fixed parameters across rounds. Adaptive schemes might enable genuine iterative improvement.

    \item \textbf{Software-dependent feature pipeline.} The wide feature projection (\S\ref{sec:pipeline}) is entirely in software, so the highest-accuracy results depend on an increasingly software-heavy pipeline.
\end{enumerate}

\section{Conclusion}
\label{sec:conclusion}

This paper reports, to our knowledge, the first federated learning experiments on physical neuromorphic hardware, totaling ${\sim}1{,}580$ trials on two Akida AKD1000 processors. Three results stand out.

First, \textbf{neuron-level aggregation (FedUnion) consistently outperforms weight averaging (FedAvg)}, across every phase, configuration, and multi-round setting we tested. The reason is that STDP neurons encode knowledge as sparse prototype vectors; averaging destroys these prototypes, while concatenation preserves them.

Second, \textbf{feature quality is the single most important lever}. Fine-tuning the upstream DS-CNN accounts for roughly nine-tenths of the accuracy gain over the baseline, and widening the feature space from 64 to 256 dimensions pushes the best federated result to $77.0\%$ ($n=30$, $p<0.001$).

Third, both the feature-width and binarization experiments point to a shared \textbf{prototype complementarity} mechanism: any change that makes per-node prototypes more distinctive amplifies the federation gain, while any change that makes them more similar (e.g., binarization, or splitting data across more nodes) reduces it.

The complete pipeline runs on commodity Raspberry Pi hardware in under three minutes per trial and exchanges only 20--40\,KB of weights per round.

\paragraph{Code availability.}
All experiment code, including the orchestrator, node workers, federation strategies, hyperparameter sweep, and analysis scripts, is publicly available at \url{https://github.com/Stemo688/federated-neuromorphic-learning}.



\begin{thebibliography}{30}

\bibitem[Bi and Poo(1998)]{bi1998synaptic}
G.-Q. Bi and M.-M. Poo.
\newblock Synaptic modifications in cultured hippocampal neurons: Dependence on spike timing, synaptic strength, and postsynaptic cell type.
\newblock \emph{Journal of Neuroscience}, 18(24):10464--10472, 1998.

\bibitem[BrainChip(2023a)]{brainchip2023}
BrainChip Holdings Ltd.
\newblock Akida neuromorphic processor: Technical reference manual.
\newblock Technical report, BrainChip, 2023a.

\bibitem[{BrainChip}(2023b)]{ds_cnn_kws}
{BrainChip}.
\newblock DS-CNN for keyword spotting: Akida models library.
\newblock \url{https://doc.brainchipinc.com/api_reference/akida_models_apis.html}, 2023b.

\bibitem[{BrainChip}(2023c)]{akida_models}
{BrainChip}.
\newblock Akida models: Pre-trained neural network models for Akida.
\newblock \url{https://github.com/Brainchip-Inc/akida_examples}, 2023c.

\bibitem[Courbariaux et~al.(2016)]{courbariaux2016binarized}
M.~Courbariaux, I.~Hubara, D.~Soudry, R.~El-Yaniv, and Y.~Bengio.
\newblock Binarized neural networks.
\newblock In \emph{NeurIPS}, 2016.

\bibitem[Cover and Hart(1967)]{cover1967nearest}
T.~Cover and P.~Hart.
\newblock Nearest neighbor pattern classification.
\newblock \emph{IEEE Transactions on Information Theory}, 13(1):21--27, 1967.

\bibitem[Davies et~al.(2018)]{davies2018loihi}
M.~Davies, N.~Srinivasa, T.-H. Lin, G.~Chinya, Y.~Cao, S.~H. Choday, G.~Dimou, P.~Joshi, N.~Imam, S.~Jain, et~al.
\newblock Loihi: A neuromorphic manycore processor with on-chip learning.
\newblock \emph{IEEE Micro}, 38(1):82--99, 2018.

\bibitem[Davis and Mermelstein(1980)]{davis1980comparison}
S.~Davis and P.~Mermelstein.
\newblock Comparison of parametric representations for monosyllabic word recognition in continuously spoken sentences.
\newblock \emph{IEEE Transactions on Acoustics, Speech, and Signal Processing}, 28(4):357--366, 1980.

\bibitem[Jacob et~al.(2018)]{jacob2018quantization}
B.~Jacob, S.~Kligys, B.~Chen, M.~Zhu, M.~Tang, A.~Howard, H.~Adam, and D.~Kalenichenko.
\newblock Quantization and training of neural networks for efficient integer-arithmetic-only inference.
\newblock In \emph{CVPR}, pages 2704--2713, 2018.

\bibitem[Kairouz et~al.(2021)]{kairouz2021advances}
P.~Kairouz, H.~B. McMahan, B.~Avent, A.~Bellet, M.~Bennis, A.~N. Bhagoji, K.~Bonawitz, Z.~Charles, G.~Cormode, R.~Cummings, et~al.
\newblock Advances and open problems in federated learning.
\newblock \emph{Foundations and Trends in Machine Learning}, 14(1--2):1--210, 2021.

\bibitem[Konecny et~al.(2016)]{konecny2016federated}
J.~Konecny, H.~B. McMahan, F.~X. Yu, P.~Richtarik, A.~T. Suresh, and D.~Bacon.
\newblock Federated learning: Strategies for improving communication efficiency.
\newblock \emph{arXiv preprint arXiv:1610.05492}, 2016.

\bibitem[Li et~al.(2020)]{li2020federated}
T.~Li, A.~K. Sahu, M.~Zaheer, M.~Sanjabi, A.~Talwalkar, and V.~Smith.
\newblock Federated optimization in heterogeneous networks.
\newblock In \emph{MLSys}, 2020.

\bibitem[Li et~al.(2023)]{li2020federated_survey}
Q.~Li, Z.~Wen, Z.~Wu, S.~Hu, N.~Wang, Y.~Li, X.~Liu, and B.~He.
\newblock A survey on federated learning systems: Vision, hype and reality for data privacy and protection.
\newblock \emph{IEEE Transactions on Knowledge and Data Engineering}, 35(4):3347--3366, 2023.

\bibitem[Maass(1997)]{maass1997networks}
W.~Maass.
\newblock Networks of spiking neurons: The third generation of neural network models.
\newblock \emph{Neural Networks}, 10(9):1659--1671, 1997.

\bibitem[McMahan et~al.(2017)]{mcmahan2017communication}
H.~B. McMahan, E.~Moore, D.~Ramage, S.~Hampson, and B.~A. y~Arcas.
\newblock Communication-efficient learning of deep networks from decentralized data.
\newblock In \emph{AISTATS}, 2017.

\bibitem[Merolla et~al.(2014)]{merolla2014truenorth}
P.~A. Merolla, J.~V. Arthur, R.~Alvarez-Icaza, A.~S. Cassidy, J.~Sawada, F.~Akopyan, B.~L. Jackson, N.~Imam, C.~Guo, Y.~Nakamura, et~al.
\newblock A million spiking-neuron integrated circuit with a scalable communication network and interface.
\newblock \emph{Science}, 345(6197):668--673, 2014.

\bibitem[Pan and Yang(2010)]{pan2010survey}
S.~J. Pan and Q.~Yang.
\newblock A survey on transfer learning.
\newblock \emph{IEEE Transactions on Knowledge and Data Engineering}, 22(10):1345--1359, 2010.

\bibitem[Pfeiffer and Pfeil(2018)]{pfeiffer2018deep}
M.~Pfeiffer and T.~Pfeil.
\newblock Deep learning with spiking neurons: Opportunities and challenges.
\newblock \emph{Frontiers in Neuroscience}, 12:774, 2018.

\bibitem[Roy et~al.(2019)]{roy2019towards}
K.~Roy, A.~Jaiswal, and P.~Panda.
\newblock Towards spike-based machine intelligence with neuromorphic computing.
\newblock \emph{Nature}, 575:607--617, 2019.

\bibitem[Shi et~al.(2016)]{shi2016edge}
W.~Shi, J.~Cao, Q.~Zhang, Y.~Li, and L.~Xu.
\newblock Edge computing: Vision and challenges.
\newblock \emph{IEEE Internet of Things Journal}, 3(5):637--646, 2016.

\bibitem[Skatchkovsky et~al.(2020)]{skatchkovsky2020federated}
N.~Skatchkovsky, H.~Jang, and O.~Simeone.
\newblock Federated neuromorphic learning of spiking neural networks for low-power edge intelligence.
\newblock In \emph{IEEE ICASSP}, 2020.

\bibitem[Snell et~al.(2017)]{snell2017prototypical}
J.~Snell, K.~Swersky, and R.~Zemel.
\newblock Prototypical networks for few-shot learning.
\newblock In \emph{NeurIPS}, 2017.

\bibitem[Venkatesha et~al.(2021)]{venkatesha2021federated}
Y.~Venkatesha, Y.~Kim, L.~Tassiulas, and P.~Panda.
\newblock Federated learning with spiking neural networks.
\newblock \emph{IEEE Transactions on Signal Processing}, 69:6183--6194, 2021.

\bibitem[Wang et~al.(2020)]{wang2020generalizing}
Y.~Wang, Q.~Yao, J.~T. Kwok, and L.~M. Ni.
\newblock Generalizing from a few examples: A survey on few-shot learning.
\newblock \emph{ACM Computing Surveys}, 53(3):1--34, 2020.

\bibitem[Warden(2018)]{warden2018speech}
P.~Warden.
\newblock Speech commands: A dataset for limited-vocabulary speech recognition.
\newblock \emph{arXiv preprint arXiv:1804.03209}, 2018.

\bibitem[Yang et~al.(2022)]{yang2022lead}
H.~Yang, K.-Y. Lam, L.~Xiao, Z.~Xiong, H.~Hu, D.~Niyato, and H.~V. Poor.
\newblock Lead federated neuromorphic learning for wireless edge artificial intelligence.
\newblock \emph{Nature Communications}, 13:4269, 2022.

\bibitem[Zhang et~al.(2017)]{zhang2017hello}
Y.~Zhang, N.~Suda, L.~Lai, and V.~Chandra.
\newblock Hello Edge: Keyword spotting on microcontrollers.
\newblock \emph{arXiv preprint arXiv:1711.07128}, 2017.

\bibitem[Zhao et~al.(2018)]{zhao2018federated}
Y.~Zhao, M.~Li, L.~Lai, N.~Suda, D.~Civin, and V.~Chandra.
\newblock Federated learning with non-IID data.
\newblock \emph{arXiv preprint arXiv:1806.00582}, 2018.

\end{thebibliography}
\end{document}